\documentclass[11pt,leqno,fleqn]{article}
\usepackage{amsthm,amsmath,amssymb}
\usepackage{graphicx}
\usepackage{subfigure}

\usepackage[authoryear]{natbib}

\bibpunct{(}{)}{;}{a}{,}{,}
\RequirePackage[OT1]{fontenc}

\theoremstyle{remark}

\theoremstyle{definition}
\newtheorem{definition}{Definition}

\newenvironment{ISItext}
{\normalsize\rm\setlength{\parindent}{1cm}\setlength{\parskip}{0pt}}
{\vskip 12pt}

\newenvironment{ISIresume}
{\normalsize\rm\setlength{\parindent}{1cm}\setlength{\parskip}{0pt}\it}
{\vskip 12pt}

\begin{document}
\newcommand{\ISItitle}[1]{\vskip 0pt\setlength{\parindent}{0cm}\Large\textbf{#1}\vskip 12pt}
\newcommand{\ISIsubtitleA}[1]{\normalsize\rm\setlength{\parindent}{0cm}\textbf{#1}\vskip 12pt}
\newcommand{\ISIsubtitleB}[1]{\normalsize\rm\setlength{\parindent}{0cm}\textbf{#1}\vskip 12pt}
\newcommand{\ISIsubtitleFig}[1]{\normalsize\rm\setlength{\parindent}{0cm}
\textbf{\textit{#1}}\vskip 12pt}
\newcommand{\ISIauthname}[1]{\normalsize\rm\setlength{\parindent}{0cm}#1 \\}
\newcommand{\ISIauthaddr}[1]{\normalsize\rm\setlength{\parindent}{0cm}\it #1 \vskip 12pt}


\ISItitle{Generalized active learning and design of statistical experiments for manifold-valued data}

\ISIauthname{Langovoy, Mikhail}
\ISIauthaddr{EPFL, Station 14, CH-1015  Lausanne, Switzerland\\
E-mail: mikhail@langovoy.com}

\ISIsubtitleB{ABSTRACT}

\begin{ISIresume}
Characterizing the appearance of real-world surfaces is a fundamental problem in multidimensional reflectometry, computer vision and computer graphics. For many applications, appearance is sufficiently well characterized by the bidirectional reflectance distribution function (BRDF). We treat BRDF measurements as samples of points from high-dimensional non-linear non-convex manifolds. BRDF manifolds form an infinite-dimensional space, but typically the available measurements are very scarce for complicated problems such as BRDF estimation. Therefore, an efficient learning strategy is crucial when performing the measurements.

In this paper, we build the foundation of a mathematical framework that allows to develop and apply new techniques within statistical design of experiments and generalized proactive learning, in order to establish more efficient sampling and measurement strategies for BRDF data manifolds.

\end{ISIresume}


\begin{ISItext}

\section{Introduction}\label{Section1}



In computer graphics and computer vision, usually either physically inspired analytic reflectance models, like \cite{Cook_Torrance_1981} or \cite{HTSG_1991}, or parametric reflectance models chosen via qualitative criteria, like \cite{Phong_1975}, or \cite{Lafortune_1997}, are used to model BRDFs. These BRDF models are only crude approximations of the reflectance of real materials. In multidimensional reflectometry, an alternative approach is usually taken. One directly measures values of the BRDF for different combinations of the incoming and outgoing angles and then fits the measured data to a selected analytic model using optimization techniques.





There were numerous efforts to use modern machine learning techniques to construct data-driven BRDF models. \cite{Brady_gen_BRDF_2014} proposed a method to generate new analytical BRDFs using a heuristic distance-based search procedure called Genetic Programming. In \cite{Brochu_2008_Active_Learning_BRDF}, an active learning algorithm using discrete perceptional data was developed and applied to learning parameters of BRDF models such as the Ashikhmin - Shirley model \cite{Ashikhmin_Shirley_2000}, while \cite{Langovoy_Numerical_Comparison_2016} treated active learning for the Cook - Torrance model \cite{Cook_Torrance_1981}. Analysis of BRDF data with statistical and machine learning methods was discussed in \cite{Langovoy_Machine_Learning_BRDF_2015},  \cite{Langovoy_BRDF_Data_Book_2015}, \cite{sole2018bidirectional}, \cite{doctor2018optimal}.

\section{Active learning and design of experiments}\label{Section2}



In general, BRDF is a 5-dimensional manifold, having 4 angular and 1 wavelength dimension. Note that even a set of 1-dimensional manifolds is infinite-dimensional (and $k$-dimensional manifolds are not to be confused with parametric $k$-dimensional families of functions). At the same time, a typical measuring device only takes between 50 and 1000 points for all the BRDF layers together. In view of this, the available measurement points are indeed very scarce for a complicated problem such as BRDF estimation. Therefore, an efficient sampling strategy is required when performing the measurements. Since sets of BRDF measurements are, in fact, observed random manifolds, we are dealing here with manifold-valued data.









Statistical design of experiments (see \cite{Fisher_1960_Design_Experiments}, \cite{Cox_Reid_2000}) is a well-developed area of quantitative data analysis. However, previous research in this field was often more concerned with (important) topics such as manipulation checks, interactions between factors, delayed effects, repeatability, among many others. This shifted the focus away from considering design of statistical experiments on structured, constrained, or infinite-dimensional data. In contrast,BRDF measurements are carried out in strictly defined settings and by qualified experts. Therefore, there is less room for human or random errors and influences. On the other hand, BRDF measurements are collections of points representing manifolds, so defining even the simplest statistical quantities in this case turns out to be a nontrivial and conceptual task.

Overall, our methodology represents a far-reaching generalization of the active machine learning framework, also generalizing the proactive learning setup of \cite{Donmez_2008_Proactive_Learning}. Active learning, as a special case of semi-supervised machine learning, oftentimes deals with finite sets of labels and aims at solving classification or clustering problems with a finite number of classes. While there have been a number of promising practical applications, most of the existing theory deals with analysis of performance of specific algorithms (query by committee, $A^2$ algorithm, or importance-weighted approach, among a few others) under rather restrictive conditions on the loss functions, incoming distributions, and other components of the learning model. For recent developments, we refer to \cite{Agarwal_2013_Paraactive_Learning},  \cite{Beygelzimer_2009_Importance}, \cite{Dasgupta_2008_Hierarchical}.






\section{Main definition}\label{Section_3}

In the most basic case, the bidirectional reflectance distribution function (BRDF), $f_r(\omega_i,\, \omega_r) )$ is a four-dimensional function that defines how light is reflected at an opaque surface. The function takes a negative incoming light direction, $\omega_i$, and outgoing direction, $\omega_r$, both defined with respect to the surface normal $\mathbf n$, and returns the ratio of reflected radiance exiting along $\omega_r$ to the irradiance incident on the surface from direction $\omega_i$. The BRDF was first defined by Nicodemus in \cite{Nicodemus1965}. The defining equation is:


\begin{equation}\label{eq1}
f_r (\omega_i,\, \omega_r) \,=\, \frac{ d \, L_r (\omega_r)}{ d \, E_i (\omega_i)} \,=\, \frac{d \, L_r (\omega_r)}{L_i(\omega_i)\cos\theta_i\, d \, \omega_i} \,.
\end{equation}

\noindent where $L$ is radiance, or power per unit solid-angle-in-the-direction-of-a-ray per unit projected-area-perpendicular-to-the-ray, $E$ is irradiance, or power per unit surface area, and $\theta_i$ is the angle between $\omega_i$ and the surface normal, $\mathbf n$. The index $i$ indicates incident light, whereas the index $r$ indicates reflected light.




Suppose we have measurements of a BRDF available for the \emph{set of incoming angles}

\begin{equation}\label{eq2}
\Omega_{inc} \,=\, \bigr\{\omega^{(p)}_i \bigr\}^{P_{inc}}_{p = 1} \,=\, \bigr\{ \bigr(\theta^{(p)}_i , \varphi^{(p)}_i \bigr) \bigr\}^{P_{inc}}_{p = 1} \,.
\end{equation}

\noindent Here $P_{inc} \geq 1$ is the total number of incoming angles where the measurements were taken. Say that for an incoming angle $\bigr\{\omega^{(p)}_i \bigr\}$ we have measurements available for angles from the \emph{set of reflection angles}

\begin{equation}\label{eq3}
\Omega_{refl} \,=\, \bigcup^{P_{inc}}_{p = 1} \Omega_{refl}(p) \,,
\end{equation}

\noindent where

\begin{equation*}
\Omega_{refl}(p) \,=\, \biggr\{\omega^{(q)}_r \biggr\}^{P_{refl}(p)}_{q = 1} \,=\, \biggr\{ \bigr(\theta^{(q)}_r , \varphi^{(q)}_r \bigr) \biggr\}^{P_{refl}(p)}_{q = 1} \,,
\end{equation*}

\noindent where $\bigr\{P_{refl}(p)\bigr\}^{P_{inc}}_{p = 1}$ are (possibly different) numbers of measurements taken for corresponding incoming angles.
Our aim is to infer the BRDF manifold (\ref{eq1}) from the above observations.

%

%
%

In general, the connection between the true BRDF and its measurements is described via a stochastic transformation $T$, i.e. $f (\omega_i,\, \omega_r) \,=\, T \bigr( f_r(\omega_i,\, \omega_r) \bigr)$, where $T \,:\, \mathcal{M} \times \mathcal{P} \times \mathcal{F}_4 \,\rightarrow\, F_4$, with $\mathcal{M} \,=\, ( M, \mathfrak{A}, \mu )$ is an (unknown) measurable space, $\mathcal{P} \,=\, ( \Pi, \mathfrak{P}, \mathbb{P} )$ is an unknown probability space, $\mathcal{F}_4$ is the space of all Helmholtz-invariant energy preserving 4-dimensional BRDFs, and $F_4$ is the set of all functions of 4 arguments on the 3-dimensional unit sphere $S^3$ in $\mathbb{R}^4$.






In order to evaluate the influence of measurement errors and to be able to measure the quality of fit of BRDF models, one needs a "measure of distance" between BRDFs. There are many choices of distances and quasi-distances available: $L_p$, $1 \leq p < + \infty$, $L_{\infty}$, \cite{Sobolev_1938} distances, Kullback-Leibler information divergence \cite{Kullback_Leibler_1951}, \cite{Mahalanobis_Distance_1936}, chi-squared distance used in correspondence analysis \cite{Langovaya_2013}. In computer science literature on BRDFs, there are few papers that study the quality of fit of BRDF models to real data. Most of these studies use the (most standard) $L_2$-norm. An alternative approach was taken in \cite{langovoy_Novel_Metric_2014}, where a perception-inspired quasi-metric for the space of BRDFs was proposed.






\section{Active manifold learning strategies}\label{Section_8}

In BRDF sampling, the equispaced-angular grid pictured in Figure \ref{figure1} is the standard. However, as was shown in \cite{Langovoy_Numerical_Comparison_2016}, this choice of measurement points leads to very inefficient sampling. Another strategy is in using uniformly distributed points on a sphere, see Figure \ref{figure2}. Since it was already understood in the community (see \cite{Hoepe_Hauer_2010}) that the standard grid is suboptimal, there were multiple heuristic attempts to propose trickier grids that better reflect the typical structure of BRDF models. A good example is shown in Figure \ref{figure3}. Ideally, the main goal of this research is to find the best sampling strategy; this strategy has to retain its optimality at least for a class of reasonable criteria, and for a sufficiently general classes of both BRDFs as well as of estimating procedures.

On the other hand, any result showing that new strategy is better than the default strategy, at least for one specific loss function, for one specific BRDF, and one specific estimating procedure, is already instrumental in understanding the general picture of learning BRDF manifolds from scarce expensive data. This basic case is straightforwardly formulated in the language of mathematical optimization, so we are able to obtain theoretical guarantees on learning accuracy, at least for some special cases. Let us outline a possible mathematical framework for BRDF sampling, in a basic case to begin with.


Consider BRDF $f \in \mathcal{F}_4$. Suppose that $f$ is measured on the finite set $\Omega_{meas}(n)$ 

\begin{equation}\label{eq40}
\Omega_{meas}(n) \,=\, \biggr\{ \bigr( \theta^{(p)}_i , \varphi^{(p)}_i, \theta^{(q)}_r , \varphi^{(q)}_r \bigr) \,\bigr|\, \bigr(\theta^{(p)}_i , \varphi^{(p)}_i \bigr) \in \Omega_{inc} \,,\, \bigr(\theta^{(q)}_r , \varphi^{(q)}_r \bigr) \in \Omega_{refl}(p) \biggr\} \,,
\end{equation}

\noindent where

\begin{equation}\label{eq42}
  n \, = \, \bigr| \Omega_{meas}(n) \bigr| \,=\, \sum_{p=1}^{P_{inc}} \bigr| \Omega_{refl}(p) \bigr| \,.
\end{equation}

\begin{definition}

Cost function $Cost$ of a measurement configuration $\Omega_{meas}$ is a Lebesgue measurable function $Cost \,:\, \mathbb{R}^{\bigr| \Omega_{meas} \bigr|} \, \rightarrow \, \mathbb{R}_{+} $.

\end{definition}


Let $Dist$ be a function (measurable for a suitably chosen $\sigma$-algebra) such that $Dist \,:\, \mathcal{F}_4 \times \mathcal{F}_4 \,\rightarrow\, \mathbb{R}_{+}$. For our purposes, we typically like $Dist$ to be inducing either a quasi-distance or a pseudo-distance on $\mathcal{F}^0_4$, where $\mathcal{F}^0_4 \, \subseteq \, \mathcal{F}_4$ is a sufficiently reach subset. As an example, a perception-based $\mu_{BRDF}$ from \cite{langovoy_Novel_Metric_2014}, was often used in our practical experiments. Standard $L_p$-distances are easier for theoretical comparisons.

\begin{definition}

Sampling strategy $\Omega$ is a sequence $\Omega = \{ \Omega_{n_0} \}_{n_0=1}^{\infty} $ where for each $n_0$ there exists an integer $n \geq n_0 $ such that $\Omega_{n_0} \,=\, \Omega_{meas}(n) $ for some measurement configuration $\Omega_{meas}(n)$ defined according to (\ref{eq40}), and for any integers $n_1 \leq n_2$ it holds that $\bigr| \Omega_{n_1} \bigr| \,\leq\,\bigr| \Omega_{n_2} \bigr|$.

\end{definition}

Consider arbitrary fixed statistical estimator of BRDFs, $\mathcal{E}_n \,:\, \mathbb{R}^n \, \rightarrow\, \mathcal{F}_4$.


\begin{definition}

Let $\Omega = \{ \Omega_{n_0} \}_{n_0=1}^{\infty} $ be a sampling strategy, and suppose that $C_{max} \,:\, \mathbb{N} \,\rightarrow\, \mathbb{R}_{+}$ be a known function. We say that the strategy $\Omega$ has uniformly admissible costs with the majorant $C_{max}$, if for all $n \geq 1$ it holds that $Cost( \Omega_n ) \,<\, C_{max} (n)$. We say that $\Omega$ has asymptotically uniformly admissible costs with the majorant $C_{max}$, if there exist $n_{min} \in \mathbb{N}$ such that for all $n \geq n_{min}$ it holds that $Cost( \Omega_n ) \,<\, C_{max} (n)$.




\end{definition}

%
%

Consider two sampling strategies: $\Omega^1$, $\Omega^2$. Suppose that both strategies have uniformly admissible costs. The problem of generalized active learning for BRDF sampling can be stated in the following way: find a sampling strategy $\Omega_{meas} \,=\, \bigr\{ \Omega_{meas}(n) \bigr\}_{n=n_{min}}^{\infty}$ such that for all $n \geq n_{min}$



\begin{equation}
\Omega_{meas}(n) \,=\, \arg\min_{\Omega \,:\, Cost(\Omega) < C_{max}} \, Dist \bigr(\, \mathcal{E}_n (f; \Omega) , f \,\bigr)
\end{equation}


\begin{definition}

Suppose $f \in \mathcal{F}_4$ is a particular (possibly unknown) BRDF. Let $\Omega^1$, $\Omega^2$ sampling strategies be sampling strategies with $C_{max}$-uniformly admissible costs. We say that strategy $ \Omega^1 $ is asymptotically more efficient for learning $f$ than the strategy $ \Omega^2 $, and write $\Omega^1 \,\succcurlyeq_{f}\, \Omega^2$, iff

\begin{equation}
  \limsup_{n \rightarrow \infty} \frac{\, Dist \bigr(\mathcal{E}_n (f; \Omega^1 (n)) , f \bigr) \,}{ \, Dist \bigr(\mathcal{E}_n (f; \Omega^2 (n)) , f \bigr) \, }  \,<1 \,.
\end{equation}

\end{definition}

Notice that, for the task of evaluating sampling quality, expected errors (over classes of BRDFs) are more interesting than maximal errors (over the same classes). Indeed, maximal errors are often dominated by degenerate counterexamples, while we are interested in a typical case behavior of our learning procedures. Therefore, we are typically interested in expected errors of the form $\mathbb{E}_{\mathcal{F}^{'}_4} \, (\, Dist (\mathcal{E}_n (f; \Omega (n)) , f) \,) \,\rightarrow\, \min \,$, where $\mathcal{F}^{'}_4 \, \subseteq \, \mathcal{F}_4$ is a sufficiently reach subset. Clearly, the choice of quasi-metric $Dist$ plays a crucial role.

%

\begin{definition}

Suppose $\mathcal{F}^{'}_4 \, \subseteq \, \mathcal{F}_4$ is a subset of the set of BRDFs. Let $\Omega^1$, $\Omega^2$ sampling strategies be sampling strategies with $C_{max}$-uniformly admissible costs. We say that strategy $ \Omega^1 $ is asymptotically more efficient for learning BRDFs of the class $\mathcal{F}^{'}_4$ than the strategy $ \Omega^2 $, and write $\Omega^1 \,\succcurlyeq\, \Omega^2$, iff

\begin{equation}
  \limsup_{n \rightarrow \infty} \frac{\, \mathbb{E}_{\mathcal{F}^{'}_4} \, \bigr(\, Dist \bigr(\mathcal{E}_n (f; \Omega^1 (n)) , f \bigr) \,\bigr) \,}{ \,  \mathbb{E}_{\mathcal{F}^{'}_4} \, \bigr(\, Dist \bigr(\mathcal{E}_n (f; \Omega^2 (n)) , f \bigr) \,\bigr) \, }  \,<1 \,.
\end{equation}

\end{definition}

\noindent Notice that this problem is neither a classification nor a regression task, as we are picking points to estimate manifolds from noisy data.

%
%
%
%
%
%
%
%
%
%
%
%
%
%
%
%


A special case of this Definition was used in \cite{Langovoy_Numerical_Comparison_2016} in order to propose more efficient BRDF sampling strategies for industrial applications.

\section{Conclusions}\label{Section_9}




BRDF manifolds form an infinite-dimensional space, but typically the available measurements are very scarce and expensive. Therefore, an efficient sampling strategy is crucial when performing the measurements. We built a mathematical framework that allows to develop and apply new techniques within statistical design of experiments and generalized proactive learning, in order to establish more efficient sampling and measurement strategies for manifold-valued BRDF data.



\end{ISItext}

%
%

%
\begin{figure}
\centering
\subfigure[]{
	\includegraphics[scale=0.25]{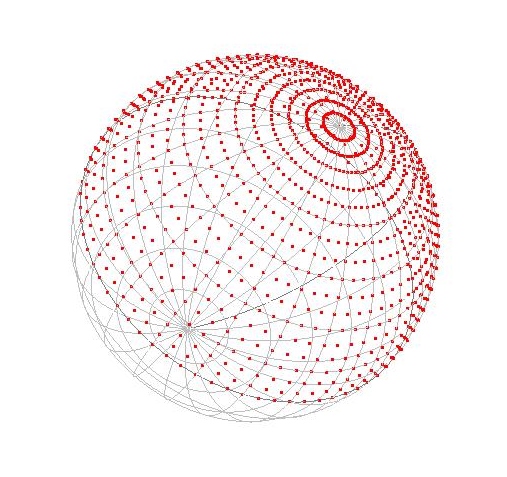}
	\label{figure1}
}
\subfigure[]{
	\includegraphics[scale=0.25]{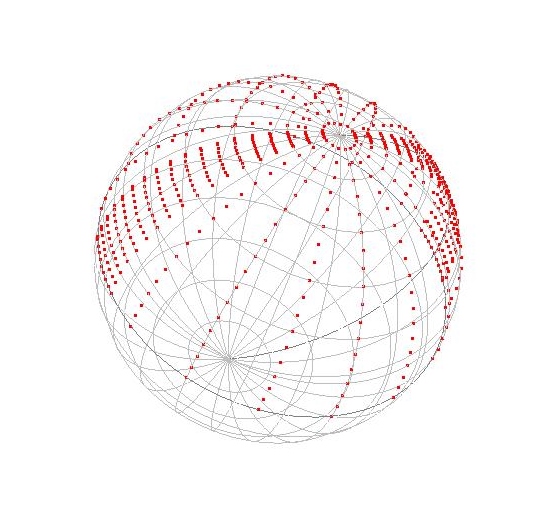}
	\label{figure3}	
}
\subfigure[]{
	\includegraphics[scale=0.25]{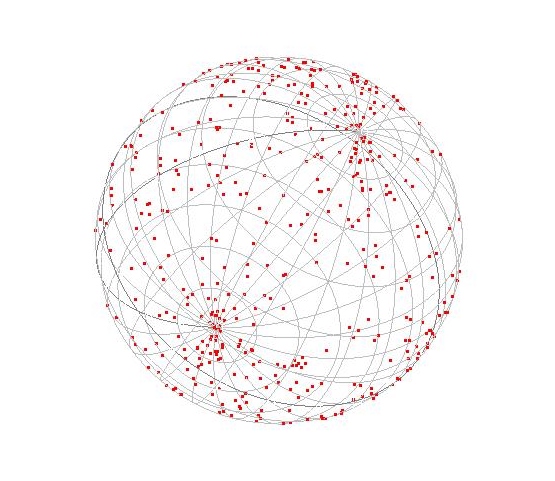}
	\label{figure2}	
}
\caption{%
Sampling strategies for BRDF manifold learning.
(a): Standard grid, inefficient sampling.
(b): Tricky grid, heuristic choice.
(c): Uniformly distributed points on a sphere.
\label{fig:cryoem}}
\end{figure}
%

%

%
%

\bibliographystyle{plainnat}
\bibliography{xDReflect_Bibliography}

%

\end{document}